\documentclass[11pt]{article}
\usepackage[a4paper,margin=2.2cm]{geometry}
\usepackage[T1]{fontenc}
\usepackage[utf8]{inputenc}
\usepackage{lmodern}
\usepackage{booktabs}
\usepackage{array}
\usepackage{microtype}
\usepackage[hidelinks]{hyperref}
\hypersetup{pdftitle={ufakzeka-kulak: Building and Evaluating a 31.6M-Parameter Turkish Speech Recognizer from Scratch},pdfauthor={Sait Furkan Teke}}
\usepackage{xurl}
\usepackage{enumitem}
\usepackage[font=small,labelfont=bf]{caption}
\setlist{nosep}
\widowpenalty=10000\clubpenalty=10000\title{ufakzeka-kulak: Building and Evaluating a 31.6M-Parameter\\ Turkish Speech Recognizer from Scratch}
\author{Sait Furkan Teke\\ ufak AI\\ \texttt{furkan@ufakai.com}}
\date{October 2026}
\begin{document}
\maketitle

\begin{abstract}
ufakzeka-kulak is an open Turkish speech-to-text model of 31.6M parameters, a FastConformer encoder with a token-and-duration transducer (TDT) and a CTC head, trained from scratch and released under Apache-2.0. It writes cased, punctuated Turkish with numbers as words. Its 4,977 training hours come from sources whose licences allow commercial use; 281 hours are human-transcribed, and for the other 4,696 the training text is an uploader subtitle or a machine transcript, kept only where a second, independent reading of the audio agrees with it. The recipe, checkpoint, labelling and export choices were made on development data, and the test sets were scored once, in a release gate fixed before training. With greedy TDT decoding the normalised word error rate (WER) is 0.0921 (95 percent interval 0.0792 to 0.1055) on FLEURS Turkish test and 0.0916 (0.0873 to 0.0958) on Common Voice 27.0 Turkish test, lower than Whisper base and Whisper small on both sets with paired intervals wholly below zero; Whisper large-v3 (greedy) scores 0.0604 and 0.0767. On conversational speech the WER is 0.3807. We also report punctuation and casing, degraded audio, and how closely each shipped export reproduces the trained model.
\end{abstract}

\section{Introduction}
Speech recognition for Turkish is available from large multilingual models such as Whisper \cite{whisper}, OWSM \cite{owsm4} and Omnilingual ASR \cite{omnilingual}. ufakzeka-kulak (\emph{kulak} is Turkish for ear) is a small model for one language, meant to run offline on a laptop, a server or in a browser page, with a licence that allows commercial use and a published list of every training source. It follows the lab's language model ufakzeka-1 \cite{ufakzeka1} and decision model ufakzeka-karar \cite{karar}.

The model is a FastConformer \cite{fastconformer} with a TDT decoder \cite{tdt} and a CTC head \cite{ctc} on the same encoder, trained from scratch with NVIDIA NeMo \cite{nemo}. We make four contributions: (i) the model, with exports for transformers \cite{transformers}, sherpa-onnx \cite{sherpa}, parakeet.cpp \cite{parakeetcpp} and parakeet-mlx \cite{parakeetmlx}, each measured against NeMo; (ii) a labelling pipeline for speech without trusted transcripts, in which one open recogniser writes the text and an independent one checks it, with both models and the agreement threshold chosen on human-transcribed speech that is not trained on; (iii) a recipe chosen by design pilots read by rules fixed before the runs, and a release gate fixed before any training; and (iv) reports beyond the gate on punctuation and casing, degraded audio, a media test set and the runtime differences we found. Every number for the released model traces to a result file committed in the project's repository. Square brackets give 95 percent intervals from a blockwise percentile bootstrap \cite{bisani} with 2,000 resamples (seed 0), in blocks of speakers for Common Voice and Real-TurnTurk, of sentences for FLEURS, which publishes no speaker ids, and of clips for ISSAI TSC dev and MediaSpeech, which have none; a difference between two systems is paired, with the same draws for both.

\section{Model}\label{sec:model}
The encoder has 16 FastConformer layers of width 256 with 4 attention heads, a feed-forward expansion of 4, a convolution kernel of 9 and depthwise-separable striding that reduces time by a factor of 8. The encoder feeds two outputs, a TDT decoder and a CTC head. The TDT decoder has a one-layer prediction network and a joint network, both of width 640, and predicts each token with a duration of 0 to 4 frames; the TDT loss uses $\sigma = 0.02$ and $\omega = 0.1$ \cite{tdt}. The CTC head is trained with weight 0.3, and an intermediate CTC loss \cite{interctc} of weight 0.3 is applied after the eighth encoder layer. Input is 16,000 Hz mono audio as 80 log-mel features with a 25 ms window and a 10 ms stride, with SpecAugment \cite{specaugment} (two frequency masks of width up to 27 and ten time masks of up to 0.05 of the utterance). The vocabulary is a SentencePiece \cite{sentencepiece} BPE of 1,024 pieces trained only on the training transcripts after decontamination. The model has 31,564,935 parameters. Unless stated otherwise, decoding is greedy, in float32, TDT with at most 10 tokens per frame.

The model writes Turkish with capital letters, the full stop, comma and question mark, the apostrophe and the hyphen inside words. Other punctuation in the training text was mapped to these marks, circumflex letters are written without the circumflex, and numbers are written as words (\emph{iki bin yirmi ikide}, not \emph{2022'de}). A training text with a number form these rules cannot read, such as a dot decimal or a Roman ordinal, was dropped; Section~\ref{sec:data} describes how ambiguous readings were resolved. It is a full-utterance model, so long recordings must be cut into utterances first.

\section{Data}\label{sec:data}
\paragraph{Sources.} Table~\ref{tab:data} lists the training pool; every source's licence allows commercial use. The human-transcribed sources are Common Voice Scripted Speech 27.0 in Turkish \cite{commonvoice}, from which every clip whose sentence or speaker is in its development or test split was removed, and Common Voice Spontaneous Speech 5.0 in Turkish; antalia-voice-corpus \cite{antalia}, one professional voice actor; and the train split of the Turkish Speech Corpus (TSC) of ISSAI \cite{issai}. ISSAI publishes the corpus under CC BY 4.0; the Hugging Face copy is tagged MIT, and CC BY 4.0 is the licence followed here. The rest of the pool is YouTube audio collected by YODAS \cite{yodas} and YODAS v3 \cite{yodas3}, whose authors say they took only videos published under a Creative Commons licence and release the audio under CC BY 3.0; on YouTube the one Creative Commons licence an uploader can choose for a video is the Creative Commons Attribution licence (CC BY) \cite{ytcc}. Its three parts are the Turkish \texttt{yodas0.10} subset of ESPnet's OWSM v4 data \cite{owsm4}, the \texttt{yodas0.00} recordings not in it, and Turkish YODAS v3 audio cut along the spans of Alania Turkish Speech Style Captions \cite{alania} (CC BY 4.0) and merged into units of at most 30 s; only the span boundaries of the Alania dataset were used, and its transcript column was never read. We call these the YODAS recordings.

\begin{table}[t]
\centering\footnotesize\setlength{\tabcolsep}{4pt}
\begin{tabular}{>{\raggedright\arraybackslash}p{6.9cm}lrrrr}
\toprule
 & & & & \multicolumn{2}{c}{hours by text origin}\\
\cmidrule(lr){5-6}
source & licence & utterances & hours & subtitle & checked\\
\midrule
Common Voice Scripted Speech 27.0 Turkish & CC0 1.0 & 64,064 & 67.90 & & \\
Common Voice Spontaneous Speech 5.0 Turkish & CC0 1.0 & 29 & 0.10 & & \\
antalia-voice-corpus & CC BY 4.0 & 892 & 4.16 & & \\
ISSAI Turkish Speech Corpus, train split & CC BY 4.0 & 178,087 & 209.06 & & \\
YODAS (\texttt{yodas0.10} and \texttt{yodas0.00}) and YODAS v3 & CC BY 3.0 & 780,714 & 4,696.21 & 369.62 & 4,326.59\\
\midrule
total & & 1,023,786 & 4,977.44 & & \\
\bottomrule
\end{tabular}
\caption{The training pool. Hours count each recording once; speed-perturbed copies and repeats of the mix are not counted. For the YODAS recordings the hours are split by the origin of the training text, the uploader's subtitle (subtitle) or a checked machine transcript (checked).}
\label{tab:data}
\end{table}

\paragraph{Text for the YODAS recordings.} Automatic captions are never training text. Whisper large-v3-turbo \cite{whisperturbo} (MIT) transcribed every unit. Where an utterance has an uploader subtitle that agrees with that transcript within a WER of 0.2, the subtitle is the training text. Otherwise, with no subtitle or one that disagrees, the training text is the transcript, kept only where Omnilingual ASR CTC 7B v2 \cite{omnilingual} (Apache-2.0), a CTC model from another maker, reads the same audio within a character error rate (CER) of 0.10; this check removed 1,134.62 hours. A number form with more than one spoken reading takes the reading that Omnilingual ASR scores highest on the audio if it beats the next by 2.0 nats; otherwise the unit keeps its rule-based reading or is dropped.

Both models and the threshold were chosen on the ISSAI TSC development split (3,428 clips, 4.26 h), which is never trained on, under rules fixed before any score was computed. Whisper large-v3 did not transcribe it significantly better than large-v3-turbo (WER 0.1573 against 0.1630, difference $-0.0057$ [$-0.0170$, $+0.0014$]), so the faster turbo was kept. Among the checking models, Omnilingual ASR CTC 7B v2 had the lowest CER, 0.0514, against 0.0599 for its 3B version and 0.0624 for Qwen3-ASR-1.7B \cite{qwen3asr} (Apache-2.0). The threshold is the loosest agreement at which the kept transcripts met fixed targets for their WER against the references and for the share of clips kept. At a CER of 0.10, 2,541 of 3,428 clips were kept, and their turbo transcripts have a WER of 0.1249, against 0.1630 for all clips. The ISSAI references are lowercase and unpunctuated and split some words, which raises every model's absolute error rate here; all models are scored against the same references. The makers of Whisper, Omnilingual ASR and Qwen3-ASR do not fully disclose their training data. No output of these models other than the transcripts and readings above is training text.

\paragraph{Decontamination.} One exclusion set holds the normalised references of every development and test set used here. It was matched against every training transcript and tokenizer line, by exact match and by shared word $n$-grams, and audio fingerprints matched the training audio against the evaluation audio. The matches removed 432 training utterances, 206 of them ISSAI TSC clips matched by their audio; near-duplicate recordings across the three YODAS parts were removed as well. FLEURS (any split), the Common Voice development and test splits, Real-TurnTurk and the 2,511 MediaSpeech clips reported here were never trained on.

\paragraph{Mix and augmentation.} The training mix samples corpora by a temperature of 0.5 with at most 4 repeats of a source per pass, so the human-transcribed sources make up 25.7 percent of the hours of a pass. The human sources are speed-perturbed \cite{speedperturb}. Noise from the MUSAN noise subset \cite{musan} (only files that MUSAN's licence files mark as public domain or CC BY) was mixed into training audio at random, reverberation was added at random with the simulated room responses of the OpenSLR room impulse response database \cite{rir} (Apache-2.0), and noise cuts with an empty transcript make up 0.8 percent of the hours of a pass as examples of non-speech. Random gain, chosen by a pilot (Section~\ref{sec:train}), multiplies half of the training cuts by a level drawn uniformly between $-40$ and $+6$ dB, after noise and room responses, so speech, noise and reverberation scale together. The cooldown mix (Section~\ref{sec:train}) raises the human share to 40 percent and draws its YODAS units only from those whose deciding agreement is at most 0.05.

\section{Training and selection}\label{sec:train}
\paragraph{Recipe.} AdamW \cite{adamw} with $\beta = (0.9, 0.98)$ and weight decay 0.001, and the Noam schedule \cite{transformer} with 8,000 warm-up steps, a peak learning rate of 0.0026 and a floor of $10^{-6}$. Training starts with a short phase on shorter clips without noise or room responses, which earlier from-scratch runs of this project needed to leave the stage in which the model outputs only blanks. The released weights are an exponential moving average of the training weights.

\paragraph{Design pilots.} Before the full run, eleven pilot runs on the full data pool, each a trunk of the full schedule plus a short cooldown, were read at step 14,500 by rules fixed before any of them started. Eight formed a $2^3$ factorial over consistency-regularised CTC \cite{crctc} against intermediate CTC, a CTC-side weight of 0.1 against 0.3, and BPE-dropout \cite{bpedropout} of 0.1 against none; two more seeds of the baseline cell measured seed noise; an eleventh run added random gain to the baseline cell. The metric was the WER pooled over the words of both development sets. The three baseline seeds scored 0.1608, 0.1583 and 0.1601, so the noise term was set by its floor of 1 percent of their mean (0.0016). A change was adopted only if it passed a Holm correction \cite{holm} over the three main effects at one-sided 0.05, lowered the metric by at least 2 percent, and its interval on FLEURS alone did not lie wholly above zero. Gain was judged on the same metric at the original level, which it had to keep within 1 percent, and on audio attenuated by 20 and 30 dB, which it had to lower by at least 5 percent and by at least 1.645 standard errors. As Table~\ref{tab:pilots} shows, none of the three recipe changes helped, and gain was adopted. The full run continued the gain pilot's trunk with the baseline recipe and gain.

\begin{table}[t]
\centering\footnotesize\setlength{\tabcolsep}{4pt}
\begin{tabular}{>{\raggedright\arraybackslash}p{5.6cm}rrll}
\toprule
compared choice & effect & relative & FLEURS-only interval & outcome\\
\midrule
CR-CTC instead of intermediate CTC & $+0.0115$ & $+7.0$\% & [$+0.0069$, $+0.0152$] & not adopted\\
CTC-side weight 0.1 instead of 0.3 & $+0.0020$ & $+1.2$\% & [$-0.0031$, $+0.0048$] & not adopted\\
BPE-dropout 0.1 instead of none & $+0.0062$ & $+3.7$\% & [$-0.0001$, $+0.0095$] & not adopted\\
random gain, original level & $+0.00005$ & $+0.03$\% & [$-0.0069$, $+0.0057$] & adopted\\
random gain, at $-20$ and $-30$ dB & $-0.0173$ & $-9.6$\% & & adopted\\
\bottomrule
\end{tabular}
\caption{Design pilots at step 14,500 on the development sets (Common Voice 27.0 dev and FLEURS dev, WER pooled over both). The effect is the mean of the runs with the choice minus the mean of the runs without it, four runs against four for the factorial rows and one run against the three baseline seeds for gain; positive is worse. The gain effect at the attenuated levels has $z = -4.19$ against the seed noise and bootstrap error.}
\label{tab:pilots}
\end{table}

\paragraph{Full run and cooldowns.} The trunk ran to step 119,500; a rule fixed in advance would have extended it had the in-training validation WER on Common Voice dev (cased and punctuated text, moving-average weights) fallen by more than 2 percent relative over the last 10,000 steps, and it had not (0.1839 to 0.1845). Two cooldowns of 16,000 steps each then ran from the same checkpoint to step 135,500, lowering the learning rate $\eta$ from its value at step 119,500, $\eta_0$, to $10^{-6}$ as $\eta_k = 10^{-6} + (\eta_0 - 10^{-6})\,(1 - \sqrt{k/K})$ after $k$ of the $K = 16{,}000$ steps \cite{cooldown}, one on the main mix and one on the cooldown mix. The candidates were each cooldown's last checkpoint and the average of its last three saved checkpoints. The rule, fixed before any was scored, took the lowest mean WER over the two development sets, sent means equal to three decimals to the candidate with fewer checkpoints, and between two such to the main mix. All four means were 0.095, so the released model is the main-mix cooldown's last checkpoint (0.0830 on Common Voice dev, 0.1071 on FLEURS dev). The two mixes do not differ significantly (cooldown-mix average minus main-mix average $-0.0009$ [$-0.0022$, $+0.0005$] on Common Voice dev and $+0.0007$ [$-0.0033$, $+0.0048$] on FLEURS dev). This training line, from the gain pilot's first step through the main-mix cooldown, ran on A100 80 GB GPUs for 52.3 GPU session hours, which count setup, validation passes and checkpoint saves as well as training steps.

\section{Evaluation}\label{sec:eval}
\paragraph{Protocol.} Every recipe, checkpoint and export choice above was made on Common Voice 27.0 Turkish dev (11,864 utterances, 12.35 h, 163 speakers) and FLEURS Turkish dev \cite{fleurs} (338 utterances, 1.12 h); the labelling choices were made on ISSAI TSC dev. The release gate was fixed before any training. The model ships only if it beats Whisper base on FLEURS Turkish test and on Common Voice 27.0 Turkish test, each by a paired interval of the WER difference that lies wholly below zero. The test sets were scored once, for one model, in a first attempt that passed. Normalised WER passes reference and hypothesis through our Turkish normaliser, which applies Unicode NFC, removes bracketed spans, writes digits as separate words with a verbaliser adapted from the TurkASR-Bench normaliser \cite{turkasr}, lowercases in the Turkish way (I to ı, İ to i), removes in-word apostrophes and all other punctuation, folds circumflexes and drops filler words such as \emph{ı}, \emph{ee}, \emph{hmm} and \emph{aa}. Whisper base, small and large-v3 \cite{whisper} (MIT) were scored by us with the same normaliser, greedy and, for base and small, with the Whisper package's decoding settings as transformers implements them (beam 5 with temperature fallback); each comparison uses the better of the two on that set. These scores were computed before any training, those on MediaSpeech before the gate.

\begin{table}[t]
\centering\scriptsize\setlength{\tabcolsep}{2.6pt}
\begin{tabular}{lrlrrll}
\toprule
 & & & & Whisper & \multicolumn{2}{c}{ufakzeka-kulak WER minus}\\
\cmidrule(lr){6-7}
set & utt. & WER & CER & base WER & Whisper base & Whisper small\\
\midrule
FLEURS test & 743 & 0.0921 [0.0792, 0.1055] & 0.0335 & 0.2461 & $-0.1541$ [$-0.1653$, $-0.1430$] & $-0.0421$ [$-0.0506$, $-0.0340$]\\
CV 27.0 test & 11,874 & 0.0916 [0.0873, 0.0958] & 0.0242 & 0.3287 & $-0.2372$ [$-0.2440$, $-0.2302$] & $-0.0885$ [$-0.0930$, $-0.0843$]\\
Real-TurnTurk & 1,532 & 0.3807 [0.3270, 0.4640] & 0.2787 & 0.7242 & $-0.3435$ [$-0.4258$, $-0.3080$] & $-0.1367$ [$-0.1544$, $-0.1164$]\\
MediaSpeech & 2,511 & 0.1042 [0.1013, 0.1073] & 0.0422 & 0.3282 & $-0.2240$ [$-0.2302$, $-0.2173$] & $-0.1248$ [$-0.1319$, $-0.1180$]\\
\bottomrule
\end{tabular}
\caption{Test results of ufakzeka-kulak, normalised WER and CER, NeMo float32, greedy TDT; CV: Common Voice; utt.: utterances (2.61, 14.54, 1.74 and 9.99 hours). Whisper base and small use the better of greedy and Whisper's own decoding per set. Only FLEURS and Common Voice gate; Real-TurnTurk has 6 speakers, so its interval is wide; MediaSpeech has no speaker ids, so its interval treats utterances as independent and may be too narrow. For reference, Whisper large-v3 (greedy) scores 0.0604 [0.0501, 0.0718] on FLEURS test, 0.0767 [0.0722, 0.0810] on Common Voice test and 0.4366 on Real-TurnTurk; no paired test was run against it.}
\label{tab:main}
\end{table}

\paragraph{Release gate.} Both comparisons pass (Table~\ref{tab:main}), and the model is also below Whisper small on both sets; Whisper large-v3 remains lower on both. The model was trained once, with one seed; the intervals come from resampling the test sets, not from repeated training. On the development sets the TDT decoder has the lower WER (Common Voice dev 0.0830 [0.0750, 0.0921] against 0.1075 [0.0983, 0.1175] for the CTC head; FLEURS dev 0.1071 against 0.1306); no paired test was run.

\paragraph{Conversational and media speech.} On Real-TurnTurk \cite{realturnturk}, conversational speech of 6 speakers, the WER is 0.3807, about four times that of read speech, and lower than Whisper base and small. MediaSpeech \cite{mediaspeech} is media speech. Its makers say part of its utterances come from Mozilla Common Voice, so every clip linked by text or audio to Common Voice or to our training audio was removed first; 2 of 2,513 clips matched our training audio, and no clip matched a Common Voice 27.0 sentence or Common Voice dev or test audio. Its references join compound numbers into one word, so part of its WER comes from the reference style; the CER with spaces removed is 0.0420. Whisper's training data is not fully disclosed, so this set may not be unseen for Whisper.

\paragraph{Punctuation and casing.} The text as written is scored on the test sets against references whose punctuation is mapped to the model's marks (Table~\ref{tab:pc}). PC-WER counts marks and capitals as words; PuncER and CaseER are the parts of the error that come from marks and from capitals \cite{cui} (the count of reference words with a capital, which that paper's equation does not define, is our reading); PER is the punctuation error rate of LibriSpeech-PC \cite{librispeechpc}, computed as NeMo implements it. Full stops are recalled far more often than commas (FLEURS test 0.839 against 0.359, Common Voice test 0.939 against 0.612). On Real-TurnTurk, where every reference ends in a mark, the model writes few marks (full-stop recall 0.171, comma recall 0.031) and often misplaces capitals (capital precision 0.365, recall 0.272).

\begin{table}[t]
\centering\footnotesize\setlength{\tabcolsep}{4pt}
\begin{tabular}{lrrrrrrr}
\toprule
 & & WER, no marks & & & & capital & references ending\\
set & PC-WER & or casing & PuncER & CaseER & PER & F1 & in a mark\\
\midrule
FLEURS test & 0.1508 & 0.0890 & 0.4223 & 0.1878 & 0.4134 & 0.8634 & 0.9960\\
Common Voice 27.0 test & 0.1132 & 0.0963 & 0.1385 & 0.0530 & 0.1379 & 0.9691 & 0.9832\\
Real-TurnTurk & 0.4960 & 0.3838 & 0.8337 & 0.3817 & 0.8647 & 0.3116 & 1.0000\\
\bottomrule
\end{tabular}
\caption{Punctuation and casing on the test sets (lower is better except capital F1). The WER without marks or casing removes only marks and capitals, with the reference numbers written as the model writes them, so it differs from the normalised WER of Table~\ref{tab:main}. The last column describes the references, not the model.}
\label{tab:pc}
\end{table}

\paragraph{Degraded audio.} FLEURS, Common Voice and Real-TurnTurk test were also scored on stored copies of their audio, made once before the gate (Table~\ref{tab:cond}); MediaSpeech has no such copies. The copies are the audio compressed with the Opus audio codec \cite{opus} at 6,000 and 12,000 bit/s in voice mode and decoded with libopus, telephone-band audio at 8,000 Hz with A-law coding, and the original audio attenuated by 20 and 30 dB. The 6,000 bit/s Opus copies cost the most and the telephone band the next most; 12,000 bit/s Opus and attenuation by 20 dB cost at most 0.0135, and at 30 dB FLEURS test loses 0.0205.

\begin{table}[t]
\centering\footnotesize\setlength{\tabcolsep}{4pt}
\begin{tabular}{llll}
\toprule
copy & FLEURS test & Common Voice 27.0 test & Real-TurnTurk\\
\midrule
Opus, 6,000 bit/s & $+0.0517$ [$+0.0458$, $+0.0579$] & $+0.0752$ [$+0.0712$, $+0.0792$] & $+0.0803$ [$+0.0604$, $+0.0920$]\\
Opus, 12,000 bit/s & $+0.0043$ [$+0.0015$, $+0.0075$] & $+0.0110$ [$+0.0090$, $+0.0130$] & $+0.0135$ [$+0.0114$, $+0.0160$]\\
telephone band, A-law & $+0.0373$ [$+0.0312$, $+0.0436$] & $+0.0491$ [$+0.0460$, $+0.0522$] & $+0.0377$ [$+0.0307$, $+0.0481$]\\
$-20$ dB & $+0.0045$ [$+0.0021$, $+0.0070$] & $+0.0009$ [$-0.0003$, $+0.0021$] & $+0.0022$ [$-0.0005$, $+0.0060$]\\
$-30$ dB & $+0.0205$ [$+0.0163$, $+0.0250$] & $+0.0051$ [$+0.0035$, $+0.0069$] & $+0.0040$ [$-0.0001$, $+0.0112$]\\
\bottomrule
\end{tabular}
\caption{WER change on degraded copies of the test audio, copy minus original, same model, paired interval.}
\label{tab:cond}
\end{table}

\section{Exports}\label{sec:exports}
Each export was decoded on both development sets and counted only if its normalised WER stayed within 0.005 of the NeMo decoding it implements on each set, with no failed utterance; a reduced setting (the int8 and q8\_0 files, and the MLX weights run at bfloat16) was also held to its full-precision counterpart by the same margin. Each runtime is compared with the NeMo decoding that follows its own rule at the per-frame token cap. Table~\ref{tab:exports} gives the files that ship.

\begin{table}[t]
\centering\footnotesize\setlength{\tabcolsep}{3pt}
\begin{tabular}{>{\raggedright\arraybackslash}p{5.0cm}lrrrr}
\toprule
 & & \multicolumn{2}{c}{utterances with NeMo's ids} & \multicolumn{2}{c}{WER minus NeMo}\\
\cmidrule(lr){3-4}\cmidrule(lr){5-6}
file & runtime & CV dev (11,864) & FLEURS dev (338) & CV dev & FLEURS dev\\
\midrule
transformers TDT, CTC & transformers & 11,864, 11,864 & 338, 338 & 0.0000 & 0.0000\\
sherpa-onnx TDT, int8, gain 8 & sherpa-onnx & 9,709 & 174 & $+0.0030$ & $+0.0022$\\
sherpa-onnx CTC, int8, gain 8, pad & sherpa-onnx & 9,339 & 142 & $+0.0032$ & $+0.0031$\\
GGUF TDT, f32 & parakeet.cpp & 11,748 & 306 & $-0.0002$ & $+0.0007$\\
GGUF TDT, q8\_0 & parakeet.cpp & 11,555 & 297 & $-0.0001$ & $+0.0018$\\
GGUF CTC, f32 & parakeet.cpp & 11,697 & 295 & $-0.0001$ & $+0.0009$\\
GGUF CTC, q8\_0 & parakeet.cpp & 11,426 & 272 & $-0.0003$ & $+0.0002$\\
MLX TDT, float32 weights, run at bfloat16, our features & parakeet-mlx & 11,744 & 331 & $+0.0001$ & 0.0000\\
\bottomrule
\end{tabular}
\caption{Shipped exports against NeMo 3.0.0 on the development sets (CV: Common Voice 27.0), with transformers 5.20.1, sherpa-onnx 1.13.8, parakeet.cpp v0.6.1, and parakeet-mlx 0.5.3 on mlx 0.32.4. The sherpa-onnx rows are the int8 files the examples load; the float32 copies also ship.}
\label{tab:exports}
\end{table}

\paragraph{transformers and parakeet.cpp.} The transformers TDT and CTC folders give NeMo's token ids on every development utterance, and the \texttt{pipeline} interface gave the same text as \texttt{generate} on 10 of 10 utterances of each set and head. One GGUF file serves both heads in parakeet.cpp; its token ids differ from NeMo's on some utterances, but WER stays within 0.0018 of NeMo.

\paragraph{sherpa-onnx front end.} The sherpa-onnx feature front end differs from NeMo's in six options. On an earlier model of the same architecture, with its own data and tokenizer, an ablation that moved each option alone found only one that closed the gap, the small constant added before the logarithm, which weighs most on low-energy frames; the others moved WER slightly. On FLEURS dev the sherpa-onnx TDT path was $+0.0108$ [$+0.0055$, $+0.0163$] against NeMo's features and $-0.0015$ [$-0.0052$, $+0.0021$] with only NeMo's constant, the change merged upstream on 21 October 2026 and released in sherpa-onnx v2.0.0 on 23 October 2026 \cite{pr3999}; the files were measured with 1.13.8. Multiplying the input by 8 before decoding is a workaround that holds at normal recording levels, and the files ship with it. For the released model the float32 files also pass at gain 1, but the int8 files fail there ($+0.0071$ for TDT and $+0.0068$ for CTC with the pad on FLEURS dev) and pass at gain 8. The gain does not help on quiet audio. At $-30$ dB the shipped TDT and CTC files lose 0.0369 and 0.0427 on FLEURS test, against 0.0205 for NeMo.

\paragraph{sherpa-onnx CTC length.} The sherpa-onnx NeMo CTC runtime sets the output length to the number of feature frames divided by the time reduction, rounded down, rather than to the encoder's own length, so the last encoder frame is dropped whenever the frame count is not a multiple of 8. In this cased model that frame often carries the final mark. Normalised WER does not see the loss. Of 11,864 Common Voice dev hypotheses, 1,788 end in a mark with the stock runtime and 11,644 with NeMo's own CTC decoding; appending zeros to the input up to the next multiple of 8 frames, which our example code does, gives 11,637. The TDT files are not affected.

\paragraph{parakeet-mlx.} parakeet-mlx computes its input features differently from NeMo, so the MLX folder ships float32 weights with a feature function that computes NeMo's features with numpy and passes them to parakeet-mlx's decoder, which runs the weights at bfloat16, its default; the package's own front end is not a checked path. bfloat16 counted only if two independent full runs gave identical token ids on every development utterance, which they did. These numbers were measured on mlx's CPU backend under Linux; the MLX files were not run on Apple silicon.

\paragraph{Shipped files on the test sets.} The int8 sherpa-onnx TDT file at gain 8 scores 0.0921 on FLEURS test and 0.0932 on Common Voice test, 0.0000 and $+0.0016$ from NeMo's TDT decoding. The int8 CTC file with the pad scores 0.1128 and 0.1200, $+0.0207$ and $+0.0284$ from NeMo's TDT decoding. That comparison includes the difference between the heads. In NeMo the CTC head's WER is 0.0245 higher than TDT's on Common Voice dev, and the CTC file is within 0.0032 of NeMo's own CTC decoding there.

\paragraph{Speed.} The shipped sherpa-onnx files were timed on 60 development utterances in three length buckets (2 to 5, 5 to 15 and 15 to 30 s), each row in a fresh process with one untimed and five timed passes. On an Apple M1 at one thread (sherpa-onnx 1.13.8, Python) the median real-time factor in the 15 to 30 s bucket is 0.0101 for the int8 TDT files and 0.0155 for the float32 ones, and the median int8 TDT decode time there is 0.170 s; in headless Chrome 155 on the same machine the demo's int8 TDT files run at 0.0311. On a 2-vCPU x86-64 Linux cloud server (Intel Xeon at 2.20 GHz, a shared virtual machine) the int8 TDT files run at a median real-time factor of 0.0364 at one thread, and every two-thread row there had a higher real-time factor than the same file at one thread. Cold load, decode times and peak memory for every file are in the model card.

\section{Limitations}\label{sec:limits}
The model reads Turkish only; other languages and Turkish mixed with them were not measured. It is not a streaming model and decodes one utterance at a time, so long recordings need voice activity detection first, with a few hundred milliseconds of audio kept around each segment; a segment cut tight to the speech can come out without capitals or punctuation. It does not separate speakers, and its word timestamps were not checked. Numbers come out as words, so an application that needs digits must convert them. Conversational speech is much harder for it than read speech, and Real-TurnTurk has only 6 speakers. Accuracy by speaker gender, age or regional accent was not measured. Commas are often missed. Most of the training text of the YODAS recordings (4,326.59 of 4,696.21 hours) was written by one recogniser and checked by another, where no uploader subtitle agreed with the transcript; errors on which the two agreed remain in the training text. The model was trained with one seed, so seed variance is not measured. The sherpa-onnx files need the input gain and, for CTC, the pad, and they lose more than NeMo on quiet audio; the MLX path was not measured on Apple silicon; transformers decodes TDT without NeMo's per-frame token cap, which changed no development utterance. MediaSpeech's interval may be too narrow, and its references join compound numbers.

\section{Release}\label{sec:release}
The Hugging Face repository holds, under Apache-2.0, the NeMo file, transformers folders for TDT and CTC, sherpa-onnx TDT and CTC files in float32 and int8, GGUF files in f32 and q8\_0, the MLX folder with float32 weights and its feature function, the Apache-2.0 licence text, one checksum file and an attribution file that credits every dataset with its licence and citation. The NeMo file is 126,955,520 bytes, the three int8 sherpa-onnx TDT files together 42,980,244 bytes and the q8\_0 GGUF file 58,147,232 bytes. The release is at these addresses.\\[2pt]
{\small\begin{tabular}{@{}ll@{}}
Weights, card and attribution & \url{https://huggingface.co/ufakai/ufakzeka-kulak}\\
Code & \url{https://github.com/ufakai/ufakzeka-kulak}\\
Demo & \url{https://kulak.ufakzeka.com}
\end{tabular}}

\paragraph{Use of AI tools.} An AI assistant was used throughout this project to write and run the data, training, evaluation and export code, to read papers and documentation, to review plans and texts, and to draft this report. The author directed the work, reviewed the results and the text, and is responsible for all of it.


{\footnotesize
\begin{thebibliography}{99}\setlength{\itemsep}{-0.8pt}\setlength{\parsep}{0pt}\raggedright
\bibitem{ufakzeka1} S. F. Teke. ufakzeka-1: Building and Evaluating a 151M-Parameter Turkish Language Model from Scratch. arXiv:2609.25081, 2026.
\bibitem{karar} S. F. Teke. ufakzeka-karar: An Open Turkish Typed-Decision Model with Order-Invariant Option Scoring. arXiv:2610.06744, 2026.
\bibitem{whisper} A. Radford et al. Robust Speech Recognition via Large-Scale Weak Supervision. arXiv:2212.04356, 2022.
\bibitem{whisperturbo} whisper-large-v3-turbo. Model card, MIT licence, \url{https://huggingface.co/openai/whisper-large-v3-turbo}.
\bibitem{owsm4} Y. Peng et al. OWSM v4: Improving Open Whisper-Style Speech Models via Data Scaling and Cleaning. Interspeech 2025. arXiv:2506.00338.
\bibitem{omnilingual} Omnilingual ASR team. Omnilingual ASR: Open-Source Multilingual Speech Recognition for 1600+ Languages. arXiv:2511.09690, 2025.
\bibitem{qwen3asr} X. Shi et al. Qwen3-ASR Technical Report. arXiv:2601.21337, 2026.
\bibitem{fastconformer} D. Rekesh et al. Fast Conformer with Linearly Scalable Attention for Efficient Speech Recognition. ASRU 2023. arXiv:2305.05084.
\bibitem{tdt} H. Xu et al. Efficient Sequence Transduction by Jointly Predicting Tokens and Durations. arXiv:2304.06795, 2023.
\bibitem{ctc} A. Graves, S. Fernández, F. Gomez and J. Schmidhuber. Connectionist Temporal Classification: Labelling Unsegmented Sequence Data with Recurrent Neural Networks. ICML 2006, pages 369 to 376. doi:10.1145/1143844.1143891.
\bibitem{interctc} J. Lee and S. Watanabe. Intermediate Loss Regularization for CTC-based Speech Recognition. ICASSP 2021. arXiv:2102.03216.
\bibitem{crctc} Z. Yao et al. CR-CTC: Consistency Regularization on CTC for Improved Speech Recognition. ICLR 2025. arXiv:2410.05101.
\bibitem{bpedropout} I. Provilkov, D. Emelianenko and E. Voita. BPE-Dropout: Simple and Effective Subword Regularization. ACL 2020. arXiv:1910.13267.
\bibitem{sentencepiece} T. Kudo and J. Richardson. SentencePiece: A Simple and Language Independent Subword Tokenizer and Detokenizer for Neural Text Processing. EMNLP 2018 (demonstrations). arXiv:1808.06226.
\bibitem{specaugment} D. S. Park et al. SpecAugment: A Simple Data Augmentation Method for Automatic Speech Recognition. Interspeech 2019, pages 2613 to 2617. arXiv:1904.08779.
\bibitem{transformer} A. Vaswani et al. Attention Is All You Need. arXiv:1706.03762, 2017.
\bibitem{adamw} I. Loshchilov and F. Hutter. Decoupled Weight Decay Regularization. ICLR 2019. arXiv:1711.05101.
\bibitem{cooldown} A. Hägele et al. Scaling Laws and Compute-Optimal Training Beyond Fixed Training Durations. NeurIPS 2024. arXiv:2405.18392.
\bibitem{holm} S. Holm. A Simple Sequentially Rejective Multiple Test Procedure. Scandinavian Journal of Statistics 6(2):65 to 70, 1979.
\bibitem{bisani} M. Bisani and H. Ney. Bootstrap Estimates for Confidence Intervals in ASR Performance Evaluation. ICASSP 2004, pages I-409 to I-412. doi:10.1109/ICASSP.2004.1326009.
\bibitem{speedperturb} T. Ko, V. Peddinti, D. Povey and S. Khudanpur. Audio Augmentation for Speech Recognition. Interspeech 2015, pages 3586 to 3589. doi:10.21437/Interspeech.2015-711.
\bibitem{rir} T. Ko, V. Peddinti, D. Povey, M. L. Seltzer and S. Khudanpur. A Study on Data Augmentation of Reverberant Speech for Robust Speech Recognition. ICASSP 2017, pages 5220 to 5224. doi:10.1109/ICASSP.2017.7953152.
\bibitem{musan} D. Snyder, G. Chen and D. Povey. MUSAN: A Music, Speech, and Noise Corpus. arXiv:1510.08484, 2015.
\bibitem{commonvoice} R. Ardila et al. Common Voice: A Massively-Multilingual Speech Corpus. LREC 2020. arXiv:1912.06670.
\bibitem{fleurs} A. Conneau et al. FLEURS: Few-shot Learning Evaluation of Universal Representations of Speech. arXiv:2205.12446, 2022.
\bibitem{issai} S. Mussakhojayeva, K. Dauletbek, R. Yeshpanov and H. A. Varol. Multilingual Speech Recognition for Turkic Languages. Information 14(2):74, 2023. doi:10.3390/info14020074. Data: \url{https://huggingface.co/datasets/issai/Turkish_Speech_Corpus}; licence CC BY 4.0, \url{https://github.com/IS2AI/TurkicASR}.
\bibitem{antalia} Antalia (Patientdesk.ai). antalia-voice-corpus. CC BY 4.0. \url{https://huggingface.co/datasets/cloud0day3/antalia-voice-corpus}.
\bibitem{yodas} X. Li et al. YODAS: Youtube-Oriented Dataset for Audio and Speech. ASRU 2023. arXiv:2406.00899.
\bibitem{yodas3} W. Chen, S. Takamichi, S. Shiota, S. Fukayama, S. Cornell and S. Watanabe. YODAS v3: Over 1 Million Hours of High-Bandwidth, Stereophonic, Multilingual Speech. Interspeech 2026, pages 5367 to 5372. \url{https://www.isca-archive.org/interspeech_2026/chen26d_interspeech.html}.
\bibitem{ytcc} YouTube Help. License types on YouTube. \url{https://support.google.com/youtube/answer/2797468}, read on 21 October 2026.
\bibitem{alania} Alania Turkish Speech Style Captions by PatientDesk AI, \url{https://huggingface.co/datasets/cloud0day3/alania-speech-captions-tr}, licensed under CC BY 4.0. Alania Turkish Speech Style Captions \copyright{} 2026 PatientDesk AI.
\bibitem{realturnturk} A. T. Bayrak, F. N. Korkmaz, B. B. Türker, M. S. Türkel and A. Kaplan. Real-TurnTurk: A Multimodal Turkish Corpus for Turn-Taking Prediction. arXiv:2608.22071, 2026.
\bibitem{mediaspeech} R. Kolobov et al. MediaSpeech: Multilanguage ASR Benchmark and Dataset. arXiv:2103.16193, 2021.
\bibitem{cui} C. Cui et al. End-to-end Joint Punctuated and Normalized ASR with a Limited Amount of Punctuated Training Data. EUSIPCO 2025. arXiv:2311.17741.
\bibitem{librispeechpc} A. Meister et al. LibriSpeech-PC: Benchmark for Evaluation of Punctuation and Capitalization Capabilities of End-to-End ASR Models. arXiv:2310.02943, 2023.
\bibitem{opus} J.-M. Valin, K. Vos and T. Terriberry. Definition of the Opus Audio Codec. RFC 6716, 2012. doi:10.17487/RFC6716.
\bibitem{turkasr} yagmurtuncer. Turkish ASR Leaderboard (TurkASR-Bench). \url{https://huggingface.co/spaces/yagmurtuncer/turkish-asr-leaderboard}. Its normaliser \texttt{kdho/normalize.py} (MIT), from which our number verbaliser is adapted, was read at commit c272516 and is no longer in the Space.
\bibitem{nemo} NVIDIA. NeMo, version 3.0.0. \url{https://github.com/NVIDIA-NeMo/Speech}.
\bibitem{transformers} T. Wolf et al. HuggingFace's Transformers: State-of-the-art Natural Language Processing. arXiv:1910.03771, 2019.
\bibitem{sherpa} k2-fsa. sherpa-onnx, version 1.13.8. \url{https://github.com/k2-fsa/sherpa-onnx}.
\bibitem{pr3999} kyo-zzz. Use NeMo's additive log-zero guard for NeMo mel features. sherpa-onnx pull request 3999, \url{https://github.com/k2-fsa/sherpa-onnx/pull/3999}, merged on 21 October 2026.
\bibitem{parakeetcpp} mudler. parakeet.cpp, tag v0.6.1. \url{https://github.com/mudler/parakeet.cpp}.
\bibitem{parakeetmlx} senstella. parakeet-mlx, version 0.5.3. \url{https://github.com/senstella/parakeet-mlx}; MLX \url{https://github.com/ml-explore/mlx}, version 0.32.4.
\end{thebibliography}}
\end{document}